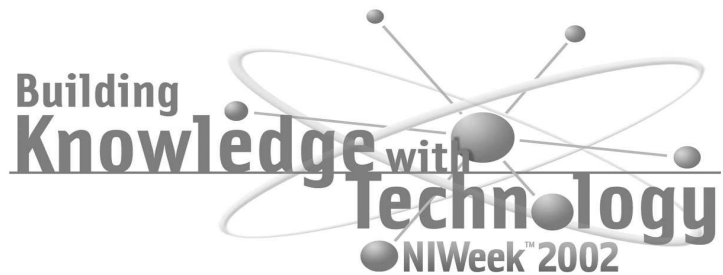

# Image Analysis in Astronomy for very large vision machine

by
Gerardo Iovane
Researcher
University of Salerno
Italy

**Category:**
R&D/Lab Automation

**Products Used:**
- LABVIEW 6i Prof Dev Sys ver 6.02
- IMAQ Vision 6.0
- SQL TOOLKIT
- INTERNET TOOLKIT
- SIGNAL PROCESSING TOOLSET

**The Challenge:**
Developing a very fast and real time system to perform image acquisition, reduction and analysis for detecting very faint luminosity variations, connected with the discovery of new planets outside the Solar System.

**The Solution:**
A real time network parallel image processing system is developed using Labview. The architecture is provided with a server, linked with a maximum of 256 workstation to analyse even 256 Mpixels images. The server takes data directly from the acquisition system, composed of a telescope with 16k×16k pixels CCD camera.

**Abstract**

It is developed a very complex system (hardware/software) to detect luminosity variations connected with the discovery of new planets outside the Solar System. Traditional imaging approaches are very demanding in terms of computing time; then, the implementation of an automatic vision and decision software architecture is presented. It allows to perform an on-line discrimination of interesting events by using two levels of triggers. A fundamental challenge was to work with very large CCD camera (even 16k×16k pixels) in line with very large telescopes. Then, the architecture can use a distributed parallel network system based on a maximum of 256 standard workstation.

# Introduction

During the last ten years much attention has been devoted to the planet detection. The passage of the planet close to the line of sight of the observer implies a luminosity variation of the star.. The presented architecture was implemented to give an answer to this challenge. There are many difficulties due to the large amount of monitored luminous objects, corresponding to never realized CCD camera (even 256 Mpixels per image). To get an effective solution, it was designed a parallel system, distributed in the world, that can analyse data and pre-processed images real time.

From a conceptual point of view the steps to process the data may be summarized as follows:
1. Measurement and image data acquisition (CCD camera and grabber are realized ad hoc);
2. Technical Image Processing (pre-reduction for bias, dark, pixel to pixel variations, cosmetics);
3. Evaluation of the quantities needed for the subsequent processing;
4. Astrometric (geometrical) alignment of images corresponding to different nights;
5. Photometric calibration of images corresponding to different nights;
6. PSF (Point Spread Function) correction of images corresponding to different nights;
7. First Trigger Level (selection of interesting luminosity variations trough a peak detection algorithm);
8. Second Trigger Level (Selection of Planet Events);
9. Pilot Analysis;
10. Off - line Analysis.

The first six points are performed by the server central unit, while the others by the client systems.

# Environment

The system architecture is structured as shown in Fig.1 while the Server interface is presented in Fig.2. All software is realized with Labview. In add the image analysis is realized with IMAQ and the data analysis use the SIGNAL PROCESSING TOOLSET by NI.

**The server machine controls the following units.**

**i)** The Image Data Acquisition (I-DAQ) Unit is responsible for the data acquisition and pre-reduction of data coming from images.

**ii)** The Control Unit, which thanks to Telescope Control System (TCS), controls the telescope by following the instructions provided by the DataBase Control System (DBCS) or by the user.

**iii)** The DataBase (DB) Unit performs the data storage and processing according to simulations or previous observations. The DB use SQL language, and is linked to the system by SQL TOOLKIT by NI.

**The clients contains the following units.**

**iv)** The Processing and Analyzing (P&A) Unit is the platform where massive data analysis is performed. It consists of three main units: 1) Data Pre-Processing Unit (DAPP) for astrometric alignment, photometric calibration, and PSF correction; Data Processing Unit (DAP) for peak detection of relevant luminosity variation, and for removing unuseful objects; Data Analysis Unit (DAU) for the fits of light curve with different expected models, color correlation, $\chi^2$ test. Step 7 of the previous paragraph is the most relevant component of the DAP Unit, and it is composed by four section: a) the peak detection procedure, b) the star detection and filtering algorithm, c) the cosmic rays filter, and d) the peak classification (single, double, multiple peak curve) methods. It is relevant to stress that for these step it can be used or the standard automatic procedures (that uses the peak detection method by NI) or the automatic unsupervised fuzzy neural network procedure (that is developed "ad hoc" with three different learning method: Multi-Layer-Self-Organizing-Maps, Neural gas, Maximum entropy). The second trigger level (step 8) tests whether the measured luminosity curves are compatible with fixed and well known models or not.

**v)** The last Unit is the so called Dispatcher which automatically builds status report about the different phases of the data flow starting from the I-DAQ to the DAU unit. In particular, statistics, plots of data and events are produced and stored by this module. Moreover, in the occurrence of a special events (like an alert or failure of the system or a short events, for which a quick answer is relevant) this unit can automatically reach and alert people with e-mail service and SMS (Short Message System). It is full developed by using the INTERNET TOOLKIT by NI.

In Fig.3 we find the client interface.

## Results of the software modules

In Fig.4 and 5 is shown a typical field takes in different conditions, while in Fig.6 we find the local corrected image (the old Fig.4) for photometric effects (see Fig.7 for histograms). Thanks to the work in Fourier transformed space all extended objects are removed (see Fig.8). Then, the light curves are built for each remaining pixel (see Figs.9).

## Conclusion

The architecture is designed for new automated planet searches performed thanks to a network distributed parallel computing. Most of the tools presented in this paper find possible applications in other fields, where pipelines and data mining procedures are called for a large amount of data. The architecture was tested on a set of data collected with the 1.3 m McGraw-Hill telescope, at MDM observatory, Kitt Peak (USA) by using the Andromeda galaxy as target. The flexibility of the system is particularly useful to select the events to study on-line and off-line and by using automatic, semiautomatic or assisted by researcher procedures. In terms of time load this procedures can produces results that cannot reach with any other traditional approach. Moreover, thanks to the real time light curves monitoring and to the dispatcher implementation, the detection of short events or events with a huge main peak and a secondary one near the first (like in planetary system) become possible. This is an other fundamental shoot reached with our system. Thanks to the presented system the Scientific Community can counts 5 very interesting events, which are compatible with new planets outside the Solar System.

## Acknowledgements

The author wish to thank A.Ambu, I.Piacentini, M.Quaglia, F.Selleri, A.Tamagnini from NI Italy for relevant software suggestions and comments.

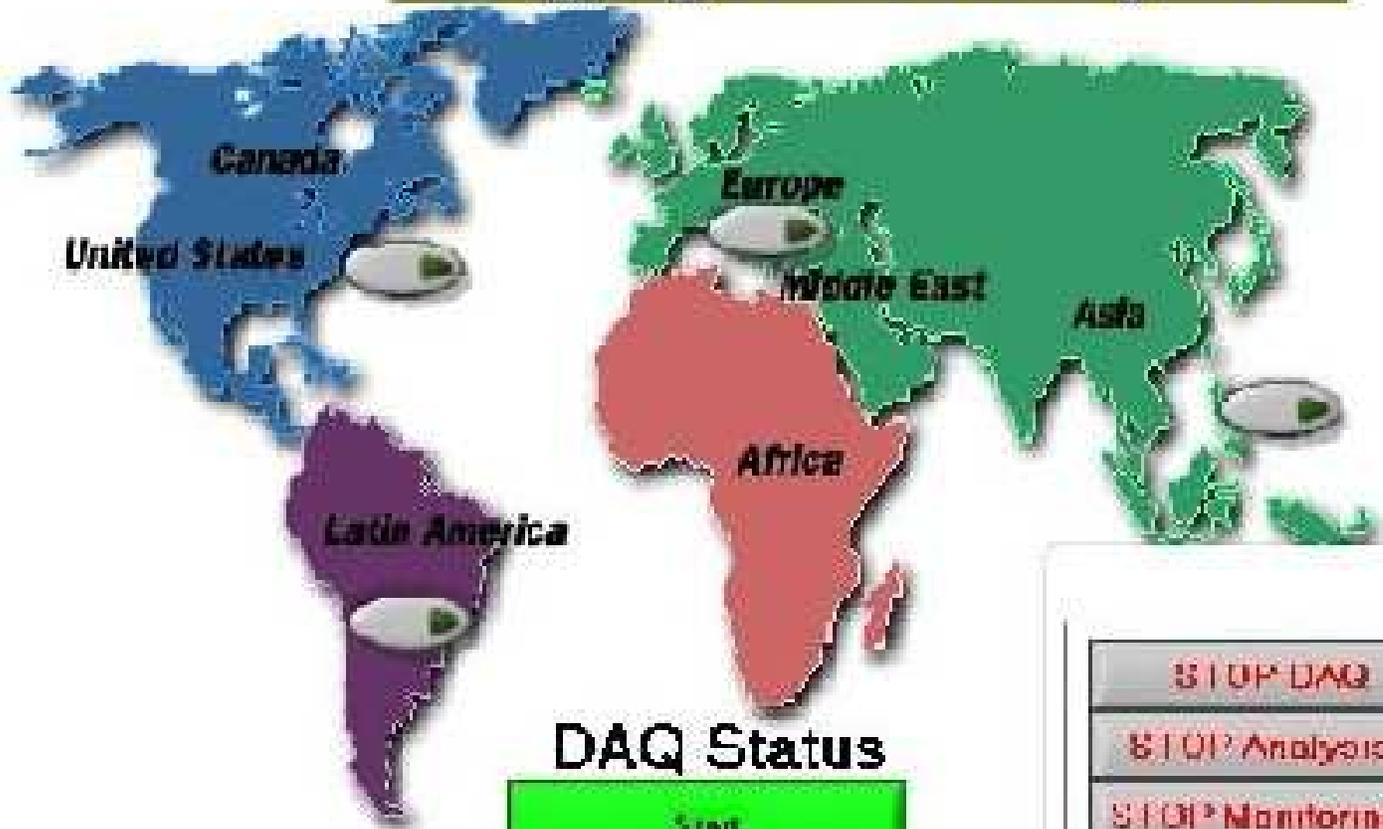

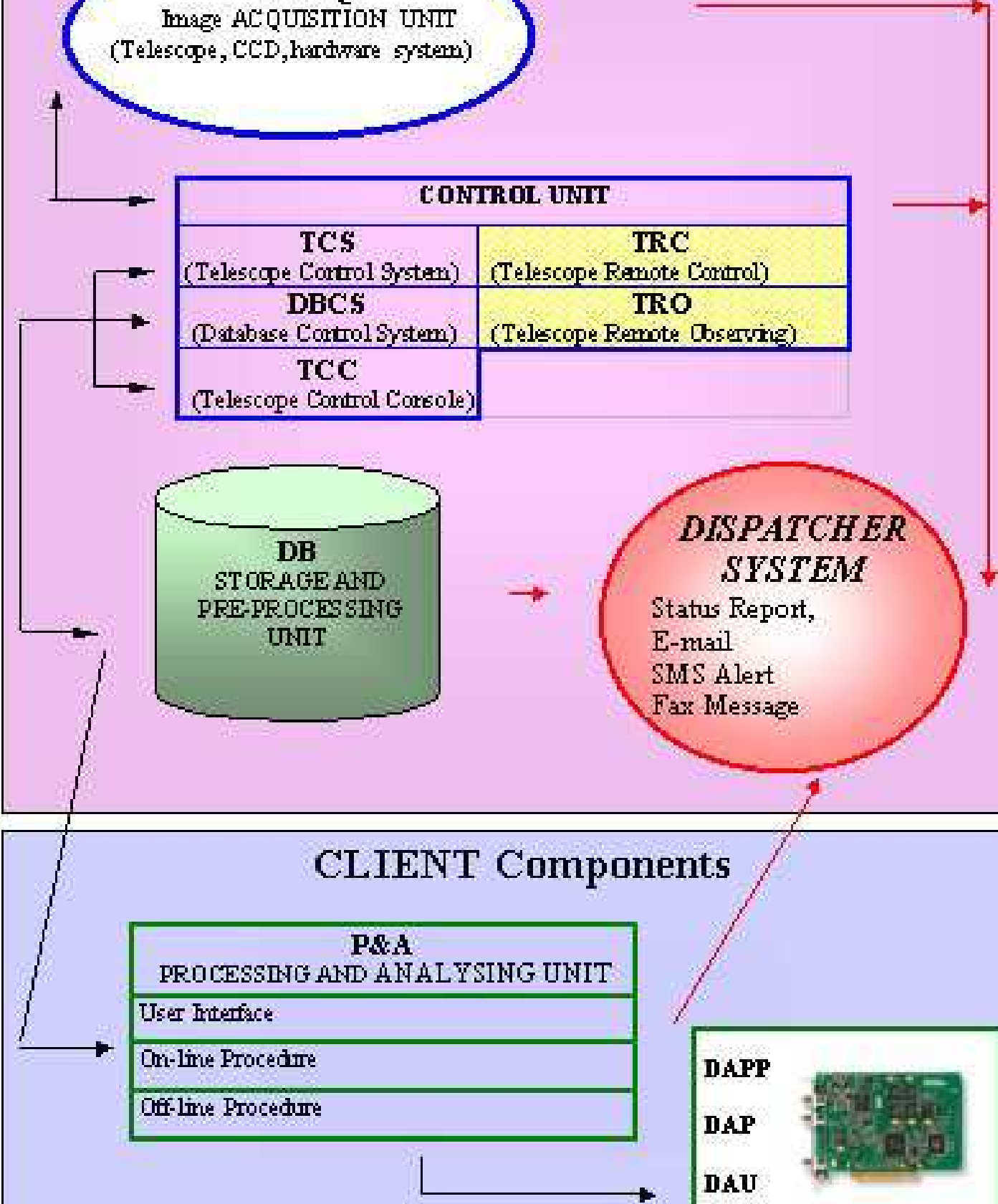

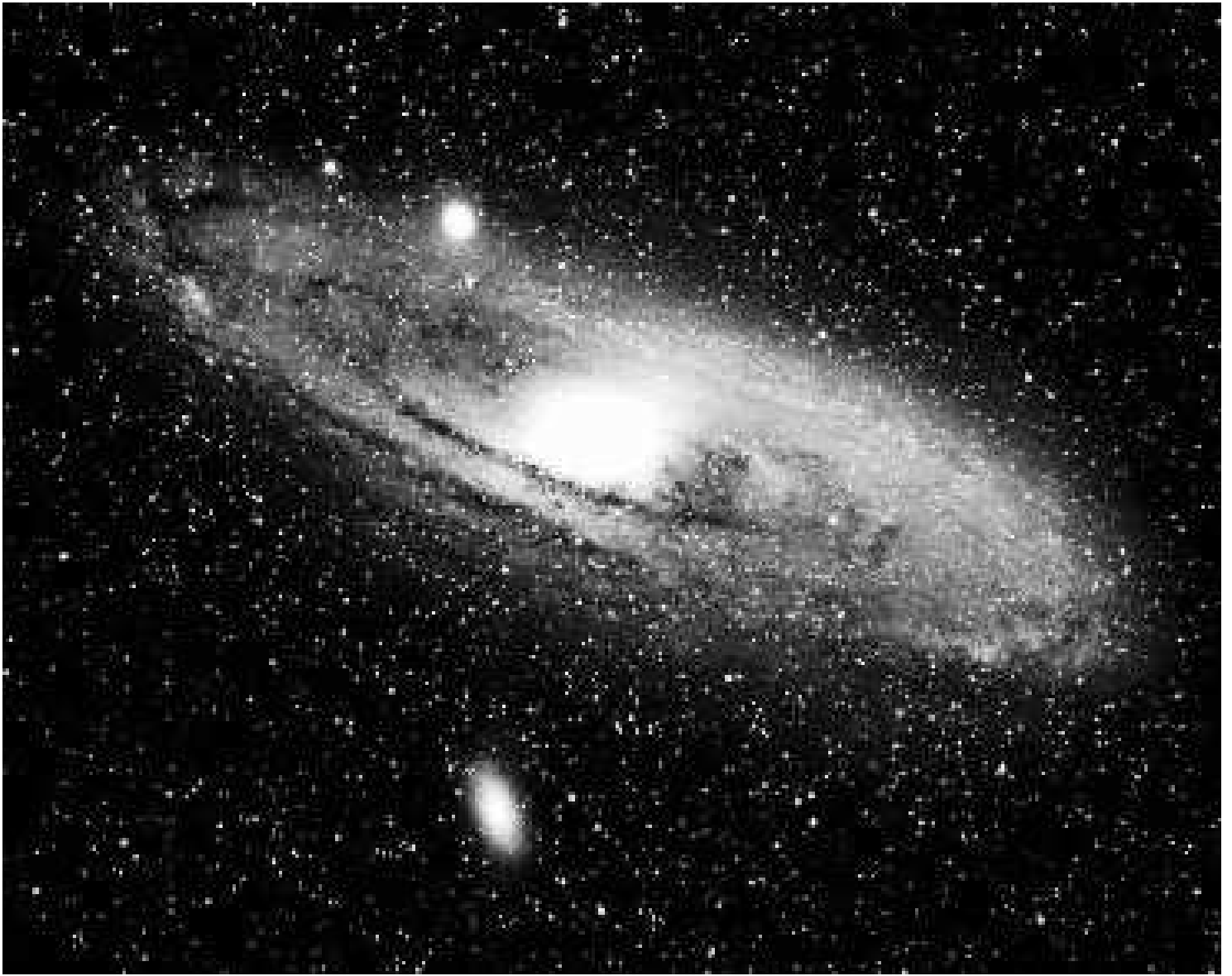

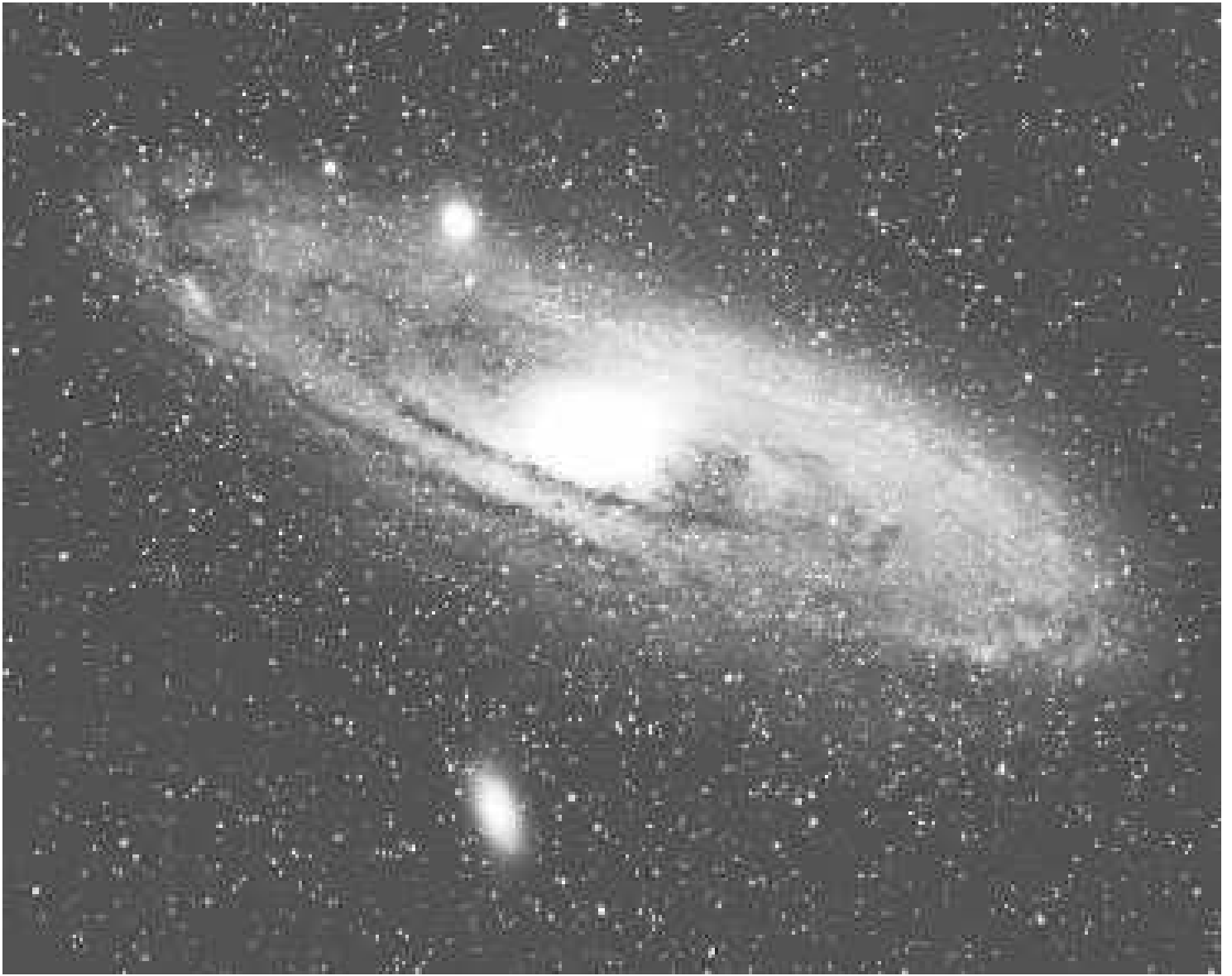

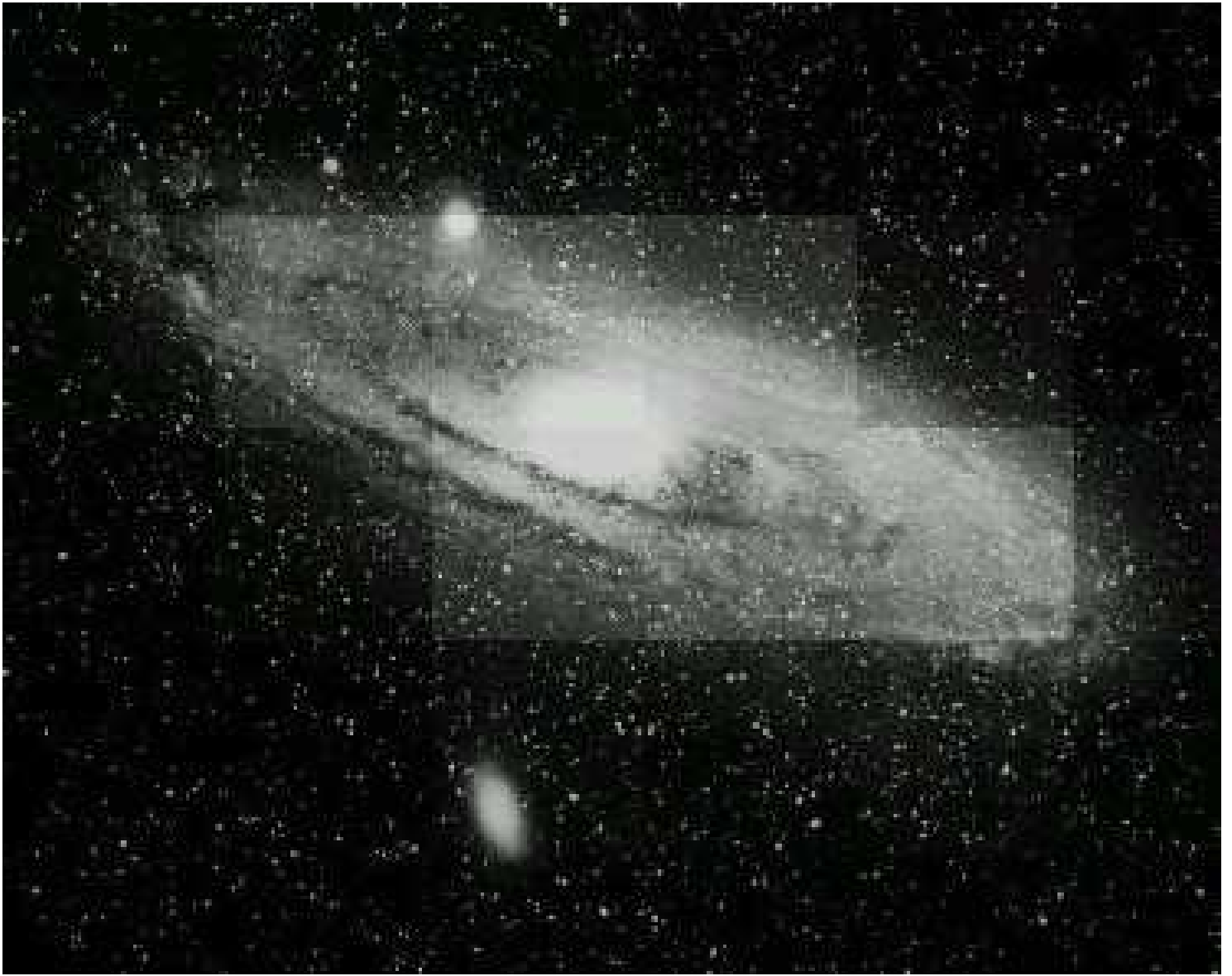

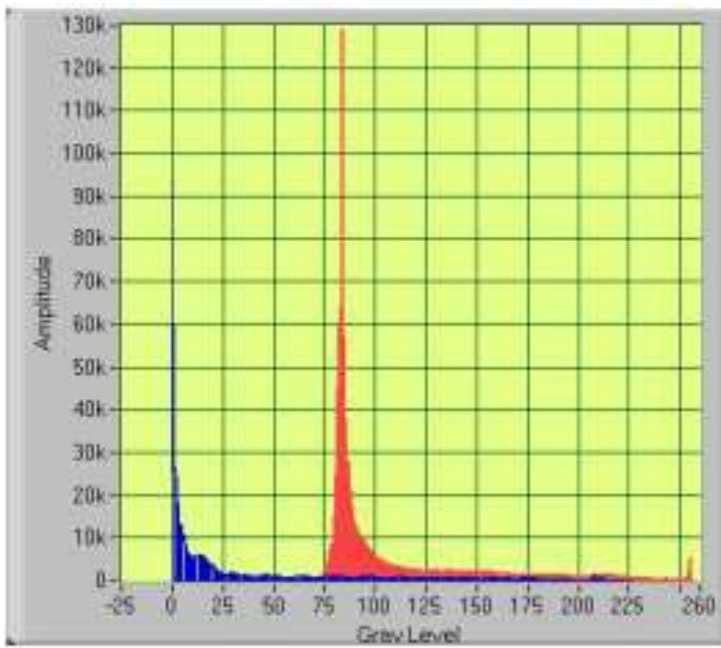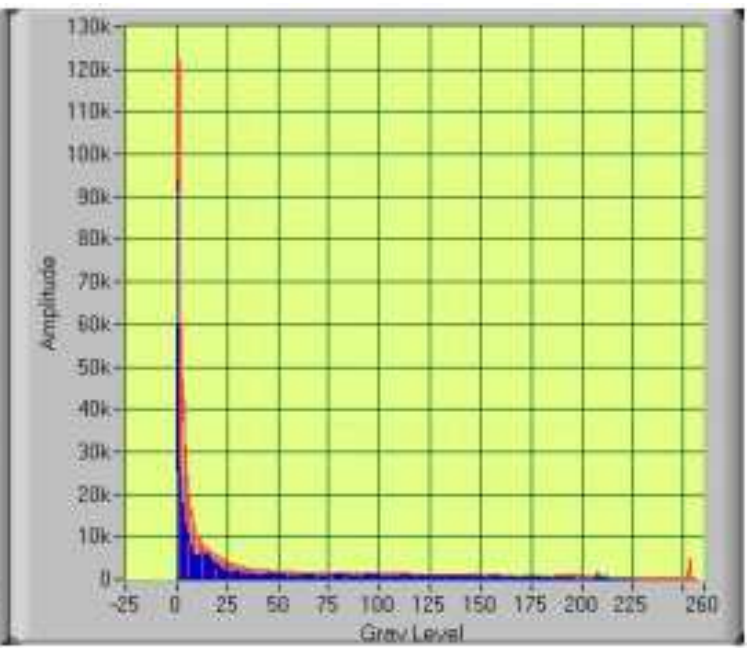

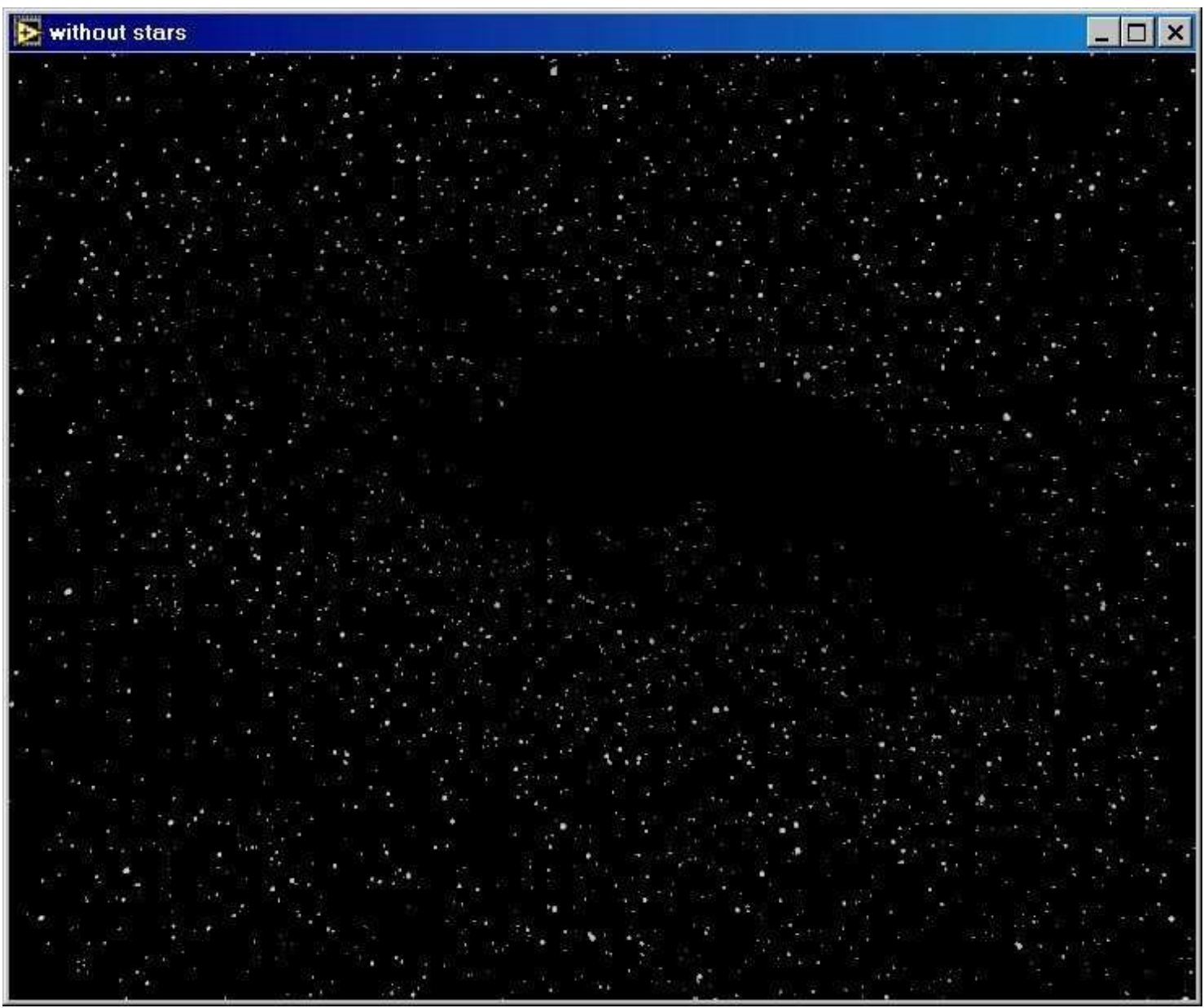

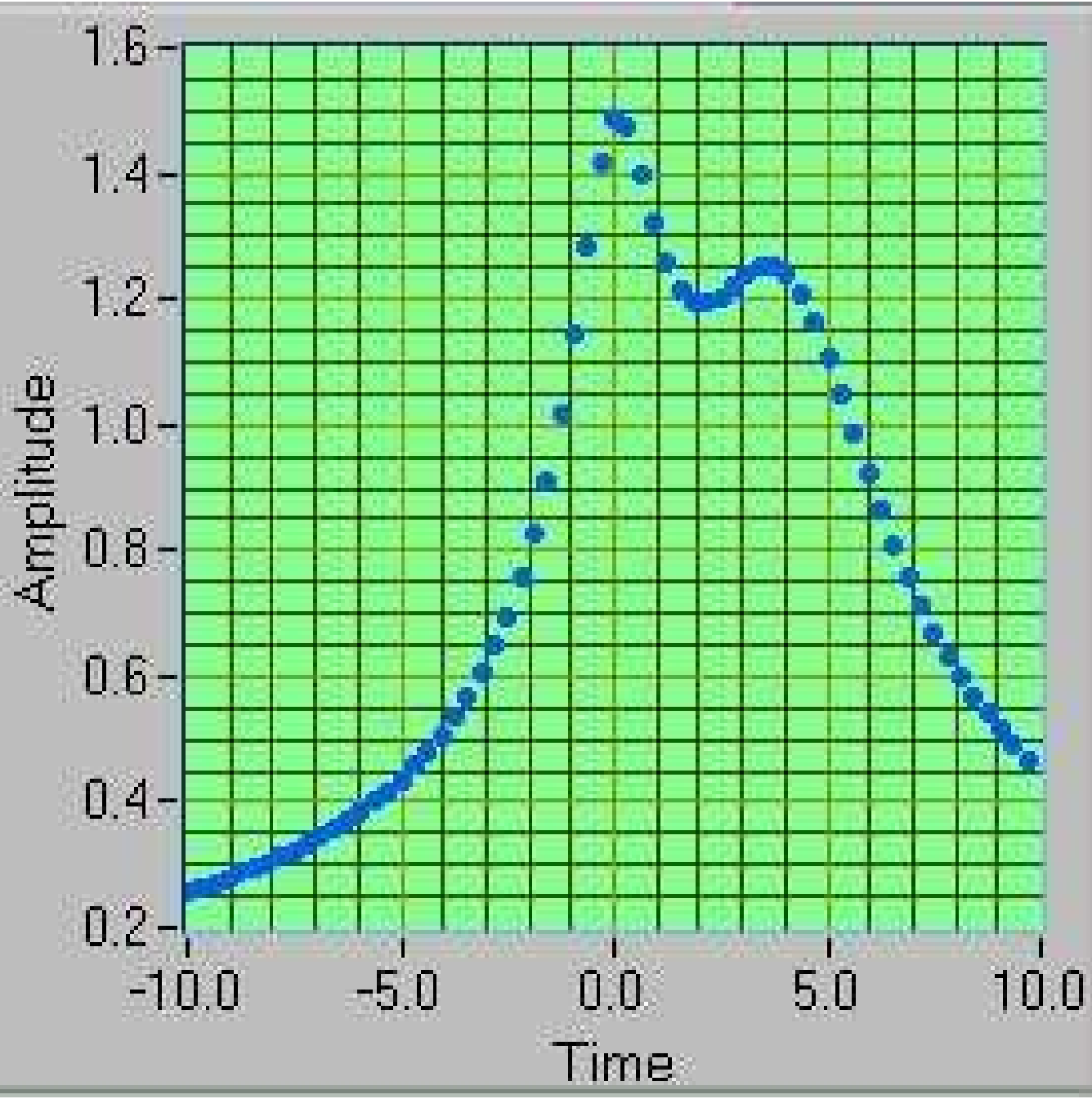